\documentclass{article} 
\usepackage{iclr2027_conference,times}
\renewcommand{\headrulewidth}{0pt}

\usepackage{amsmath,amsfonts,bm}

\def\eqref#1{equation~\ref{#1}}

\def\1{\bm{1}}

\DeclareMathAlphabet{\mathsfit}{\encodingdefault}{\sfdefault}{m}{sl}
\SetMathAlphabet{\mathsfit}{bold}{\encodingdefault}{\sfdefault}{bx}{n}

\definecolor{citecolor}{HTML}{2980b9}
\definecolor{linkcolor}{HTML}{c0392b}
\usepackage[pagebackref=true,breaklinks=true,colorlinks,bookmarks=false,citecolor=citecolor,linkcolor=linkcolor]{hyperref}
\usepackage{url}
\usepackage{xcolor}
\usepackage{booktabs}
\usepackage{multirow}
\usepackage{amsmath}
\usepackage{graphicx}
\usepackage{pifont}
\usepackage{algorithm}
\usepackage{algpseudocode}
\usepackage{colortbl}

\definecolor{bestcolor}{HTML}{FFF65B}
\definecolor{secondcolor}{HTML}{FFFBCB}

\newcommand{\cmark}{\textcolor{green!60!black}{\ding{51}}}
\newcommand{\xmark}{\textcolor{red!75!black}{\ding{55}}}

\newcommand{\fade}[1]{\textcolor{black!60}{#1}}

\title{Watch, Infer, Coordinate: Inferring Robot Partner Constraints for Zero-Shot Coordination}

\author{%
Suyu Ye$^{1,2\dagger}$ \quad Zheyuan Zhang$^{1,2\dagger}$ \quad Vaishnav Tadiparthi$^{1}$ \quad Hossein Nourkhiz Mahjoub$^{1}$ \\
\textbf{Ehsan Moradi Pari$^{1}$ \quad Tianmin Shu$^{2}$ \quad Homanga Bharadhwaj$^{2}$ \quad Nakul Agarwal$^{1}$} \\
$^{1}$Honda Research Institute USA \quad $^{2}$Johns Hopkins University
}

\iclrfinalcopy

\newcommand{\ours}{Watch, Infer, Coordinate}

\begin{document}
\vspace*{-50pt}

\maketitle
{\renewcommand{\thefootnote}{\fnsymbol{footnote}}%
\footnotetext[2]{Work done during an internship at Honda Research Institute USA.}}
{\renewcommand{\thefootnote}{}%
\footnotetext{Project page: \url{https://watch-infer-coordinate.github.io/}}}

\begin{abstract}
Robots operating in the physical world will increasingly need to coordinate with other robots, particularly in manipulation tasks where an object may be too large or heavy for a single robot to carry alone. Physical limitations caused by hardware degradation or actuator faults can restrict the actions a robot can reliably execute, yet these limitations may be unknown to its partner. We study whether a helper can infer a robot partner's physical constraints from observing it coordinate with another robot, then use the inferred capability to coordinate with the same partner on a new task. This is difficult because a demonstration shows what the constrained robot did, but not what it could have done. In physically coupled tasks, the other robot may also compensate for its limitations, making those limitations difficult to identify from the constrained robot's behavior alone. Our key insight is that these constraints shape the joint behavior of the team, making the actions of both robots informative about the constrained partner’s capability. We introduce \ours{}, a benchmark spanning three physically coupled manipulation settings, together with an inference approach that scores candidate constraints using observed joint behavior. Across all three settings, our method substantially improves constraint inference and zero-shot coordination, approaching an oracle with access to the true constraints.
\end{abstract}

\begin{figure*}[h!]
    \centering
    \includegraphics[width=0.9\textwidth, trim={1cm 5.8cm 3.5cm 0}, clip]{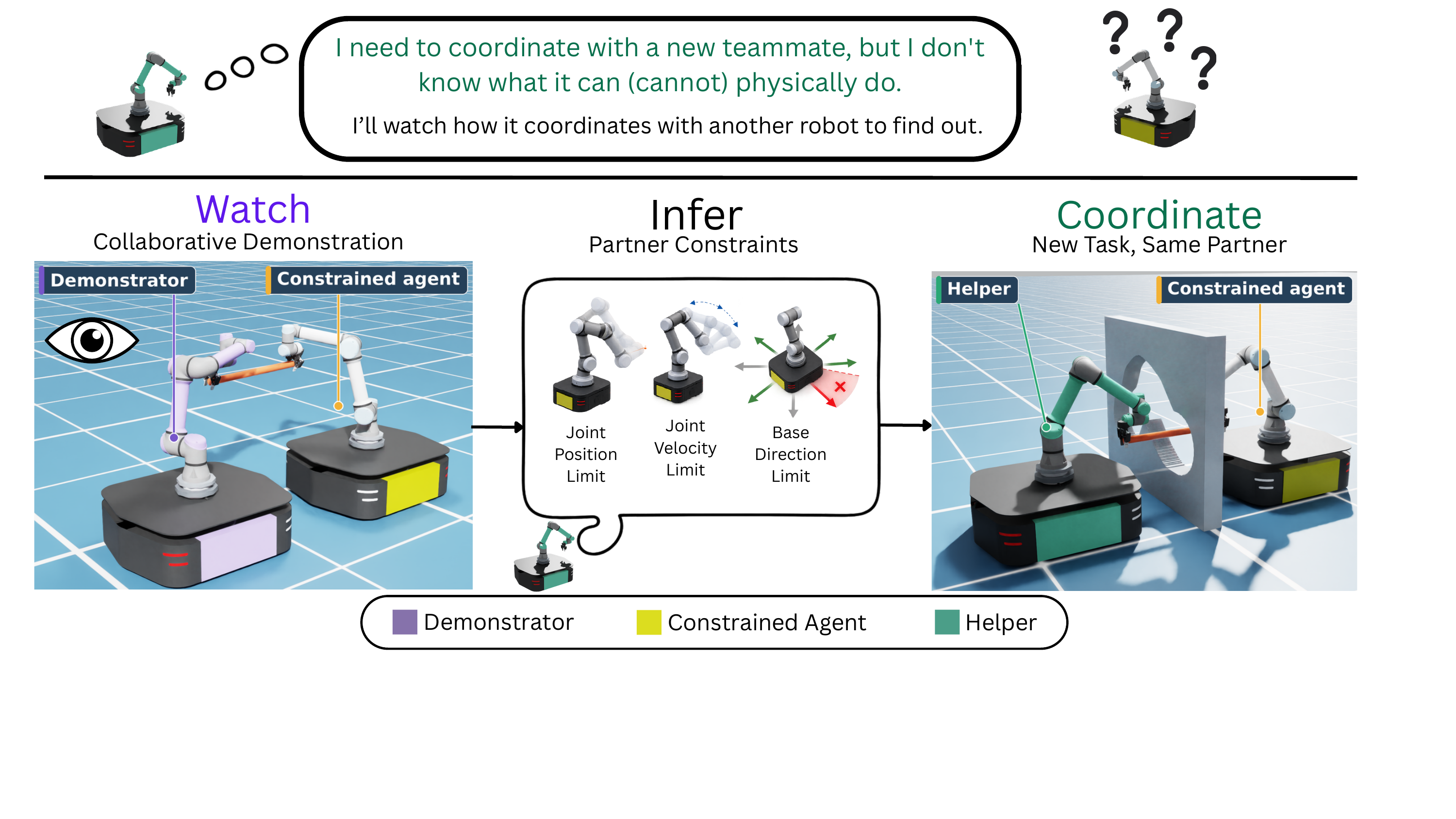}
    \caption{
\textbf{Overview of our problem setting.} A helper must coordinate with a constrained partner whose physical capabilities are initially unknown. Rather than interacting with the partner directly, the helper first watches it coordinate with another robot and uses this demonstration to infer constraints on its low-level actions, such as joint position, joint velocity, or base-direction limits. The helper then replaces the demonstrator and must coordinate with the same constrained partner on a new task. Although the required coordination changes between tasks, the partner's physical constraints remain the same, allowing the helper to transfer what it learned about the partner's capabilities to the new task.
    }
    \label{fig:overview}
    \vspace{-5pt}
\end{figure*}

\section{Introduction}

As robots move from controlled laboratories into the physical world, they will increasingly need to not only adapt to their environment, but also to learn to coordinate with one another. Some manipulation tasks naturally require multiple robots, for example when transporting objects that are too large or heavy for a single robot to handle alone \citep{farivarnejad2022multirobot}. In such physically coupled tasks, the robots' actions jointly determine the motion of the shared object, requiring closely coordinated movements, often in decentralized settings in which robots cannot access one another’s plans \citep{pandit2025multi, doshi2026chorus, yang2025implicit, shi2026safe}. In practice, a robot's physical capabilities may also change over time due to hardware degradation or actuator faults. Such failures can reduce actuator effectiveness or restrict the motion of individual joints, changing the set of actions the robot can reliably execute \citep{lewis1997fault, tinos2007fault}. In this work, we consider a helper that must coordinate with a robot, which we call the \emph{constrained agent}, subject to unknown constraints on its low-level physical actions, such as position or velocity limits on individual joints.

Existing methods do not directly address how a helper could learn these constraints from observing prior coordination. Partner-aware assistance approaches have inferred properties such as goals or task-level capabilities \citep{puig2021watchandhelp, cao2024smart, du2024constrained}, while methods for physically coupled coordination typically adapt to a partner's behavior without explicitly inferring persistent constraints on the actions that partner can execute \citep{yang2025implicit, pandit2025multi, doshi2026chorus}. We therefore ask whether a helper can infer a partner's physical constraints by observing it coordinate with another robot, and then use that knowledge to coordinate with the same partner on a new task. As illustrated in Figure~\ref{fig:overview}, the helper first observes a demonstrator coordinate with the constrained agent, then takes the demonstrator's place on a different task. Because the helper has not observed how the constrained agent coordinates on this new task before deployment, we refer to this as \emph{zero-shot coordination}.

Inferring these constraints from watching prior coordination is difficult because a demonstration shows what the constrained agent did, but not what it could have done. If the constrained agent does not take a particular action, that action may be physically infeasible or simply unnecessary for the task. Physical coupling makes this ambiguity more subtle, as the demonstrator may compensate for the constrained agent's limitations. This can make the constraint difficult to identify from the task outcome or the constrained agent's actions alone. Our key insight is that a partner's physical constraints shape the joint behavior of the team, so the actions of both agents can be used to reason about them. Because these low-level constraints belong to the agent rather than a particular task, they can also provide a capability model that remains useful when the task changes. We study these ideas in simulation settings with varying levels of complexity, allowing us to systematically vary physical constraints across different coordination problems.

We introduce \textbf{\ours}, a framework for studying coordination under unknown partner constraints. As shown in Figure~\ref{fig:tasks}, \ours{} includes a benchmark spanning three joint-carrying settings of increasing physical complexity, with obstacle-free demonstrations followed by a new task at test time in which obstacles block the direct path to the goal and require different coordinated motions. The benchmark includes 2D rod carrying, fixed-base dual-UR5 carrying, and mobile dual-UR5 carrying, spanning a low-dimensional abstraction to increasingly complex robotic manipulation. We also propose an inference approach that uses the observed actions of both agents to determine which candidate constraints are most consistent with their coordination. We use the resulting capability estimate to support zero-shot coordination with the constrained agent on the new task.

To summarize, our main contributions are fourfold. (1) We formulate the problem of inferring a robot partner's persistent low-level physical constraints from prior multi-agent coordination and using them for zero-shot coordination on a new task. (2) We introduce the \ours{} benchmark, spanning three joint-carrying settings of increasing physical complexity. (3) We propose an inference approach that uses the actions of both agents to infer the constrained agent's physical constraints from observed coordination. (4) We show that our approach improves both constraint inference and downstream task success across all three settings, approaching the performance of an oracle that is given the true constraints.

\section{Related Works}

\begin{table}[h]
\centering
\caption{
Comparison with related work.
Prior-Trajectory Inference indicates that episode-specific information is inferred from trajectories observed before the downstream task.
Multi-Agent Trajectories indicates that the trajectories used for inference contain behavior from multiple interacting agents.
Low-Level Partner Constraints indicates that the inferred quantity specifies low-level constraints such as limits on the partner’s joint positions, joint velocities, or allowable motion directions, rather than task-level capabilities such as reachable height or liftable weight.
Continuous Control indicates continuous-valued physical control rather than symbolic or macro actions.
Physically Coupled Manipulation indicates that multiple agents jointly manipulate the same object such that their actions mechanically combine to determine its motion.
Transfer to New Task indicates that the inferred information is subsequently used on a distinct task or goal requiring different behavior.
}
\label{tab:related_work_comparison}

\setlength{\tabcolsep}{3pt}
\renewcommand{\arraystretch}{1.1}

\resizebox{\linewidth}{!}{%
\begin{tabular}{@{}lcccccc@{}}
\toprule
Work
& \shortstack{Prior-Trajectory\\Inference}
& \shortstack{Multi-Agent\\Trajectories}
& \shortstack{Low-Level Partner\\Constraints}
& \shortstack{Continuous\\Control}
& \shortstack{Physically Coupled\\Manipulation}
& \shortstack{Transfer to\\New Task} \\
\midrule

CHORUS \citep{doshi2026chorus}
& \xmark & \xmark & \xmark & \cmark & \cmark & \xmark \\

Watch-and-Help \citep{puig2021watchandhelp}
& \cmark & \xmark & \xmark & \xmark & \xmark & \xmark \\

TICC-POMDP \citep{lee2020getting}
& \xmark & \xmark & \xmark & \xmark & \xmark & \cmark \\

Smart Help \citep{cao2024smart}
& \xmark & \xmark & \xmark & \xmark & \xmark & \xmark \\

CHAIC \citep{du2024constrained}
& \xmark & \xmark & \xmark & \xmark & \cmark & \xmark \\

CE-CM-Div \citep{tisnikar2026partner}
& \cmark & \cmark & \xmark & \xmark & \xmark & \cmark \\

CAMO \citep{tisnikar2024probabilistic}
& \cmark & \xmark & \cmark & \cmark & \xmark & \cmark \\

GRACE \citep{liu2025grace}
& \xmark & \xmark & \cmark & \cmark & \xmark & \cmark \\

ICRL Benchmark \citep{liu2022benchmarking}
& \cmark & \xmark & \xmark & \cmark & \xmark & \xmark \\

MACL \citep{zhang2026constraint}
& \cmark & \cmark & \xmark & \cmark & \xmark & \cmark \\

\midrule
\textbf{Ours}
& \cmark & \cmark & \cmark & \cmark & \cmark & \cmark \\

\bottomrule
\end{tabular}%
}
\end{table}

\textbf{Physically Coupled Robot Collaboration.} Prior work has studied physically coupled robot collaboration in both human-robot and robot-robot teams. Human-robot methods adapt to the partner's current behavior, for example by inferring intent from joint motion \citep{yang2025implicit}, using language to revise motion plans \citep{shi2026safe}, or learning cooperative policies from human-human demonstrations \citep{ng2023diffusion}. Multi-robot methods learn decentralized controllers using shared \citep{pandit2024learning,pandit2025multi,doshi2026chorus}, or agent-specific policies \citep{he2025latent, dong2025mimic}. These methods span model-based planning \citep{stouraitis2020online,kennel2024payload, jaafar2024mr, yang2025implicit}, reinforcement learning \citep{nachum2019multi, zhang2020decentralized, shibata2023deep, aljalbout2023clas, pandit2024learning, pandit2025multi}, diffusion policies \citep{dong2025mimic, ng2023diffusion, he2025latent}, and VLM/VLA-based policies \citep{song2025collabot, doshi2026chorus}. However, they do not explicitly infer a persistent physical constraint on the actions that a particular partner can execute. Ability-aware controllers can account for differences in robots' actuation limits, but assume those limits are known \citep{yan2021decentralized}. Our work instead infers the constrained agent's feasible action set from prior physically coupled demonstrations and uses the inferred model to coordinate with the same agent in a new environment.

\textbf{Partner-Aware Assistance.}
The watch-and-help challenge proposed the problem of observing an agent, inferring its goal, and then helping it achieve that goal \citep{puig2021watchandhelp}. Subsequent work has developed methods for inferring and updating the agent's goal online \citep{puig2023nopa, zhi2020online}. More broadly, partner-aware assistance methods has modeled other properties that shape an agent's behavior, including its beliefs about environment dynamics \citep{reddy2018you}, and its latent strategy across repeated interactions \citep{xie2021learning}.  Capability-aware methods instead reason about what the partner can do, but often represent capability through task-level properties or high-level actions. TICC-POMDP models capability through action success probabilities \citep{lee2020getting}. Smart Help estimates semantic capabilities such as reachable height and liftable weight \citep{cao2024smart}, while CHAIC  uses VLM and LLM based helpers to infer a partner's goals and physical limitations online and reasons about these limitations through task-level actions and assistance plans \citep{du2024constrained}. CE-CM-Div infers domain-specific capabilities defined over symbolic and macro actions \citep{tisnikar2026partner}. Our work instead infers a partner's low-level physical action constraints, such as joint and velocity limits, from demonstrations of the partner collaborating with another agent. A new helper then uses this capability model to plan joint motion with the partner, zero-shot, on a new task.

\textbf{Constraint Inference from Demonstrations.} The goal of constraint inference from demonstrations is to recover unknown constraints on an agent or its environment from observed behavior \citep{chou2018learning,scobee2019maximum,malik2021inverse}. Optimization-based methods search for lower-cost alternatives to the demonstrated trajectory, relying on the idea that if the demonstration is optimal, these alternatives must violate an unknown constraint \citep{chou2018learning, chou2020learning}. Other methods infer constraints by finding those that make the observed demonstrations most likely \citep{scobee2019maximum,malik2021inverse}. However, much of this work focuses on environmental constraints, such as hidden obstacles and forbidden regions, that are directly reflected in a single agent's trajectory. Prior work on agent centric constraint inference either observes the agent acting on its own \citep{tisnikar2024probabilistic, liu2025grace} or models capability as domain-specific restrictions on which tools or task-level actions the agent can use \citep{tisnikar2026partner}. Recent work has also inferred constraints from multi-agent demonstrations, but focuses on interaction constraints between agents rather than the physical limits of a particular partner \citep{zhang2026constraint}. Our work instead infers a partner's low-level physical limits from cooperative, physically coupled team demonstrations, where the helper may compensate for the constrained partner, and the resulting team motion can hide the partner's limitations.

\section{\ours{}: Problem Setup and Benchmark}

\begin{figure*}[t]
    \centering
    \vspace{-5pt}
    \includegraphics[width=\textwidth, trim={1.5cm 7.5cm 14.5cm 0cm}, clip]{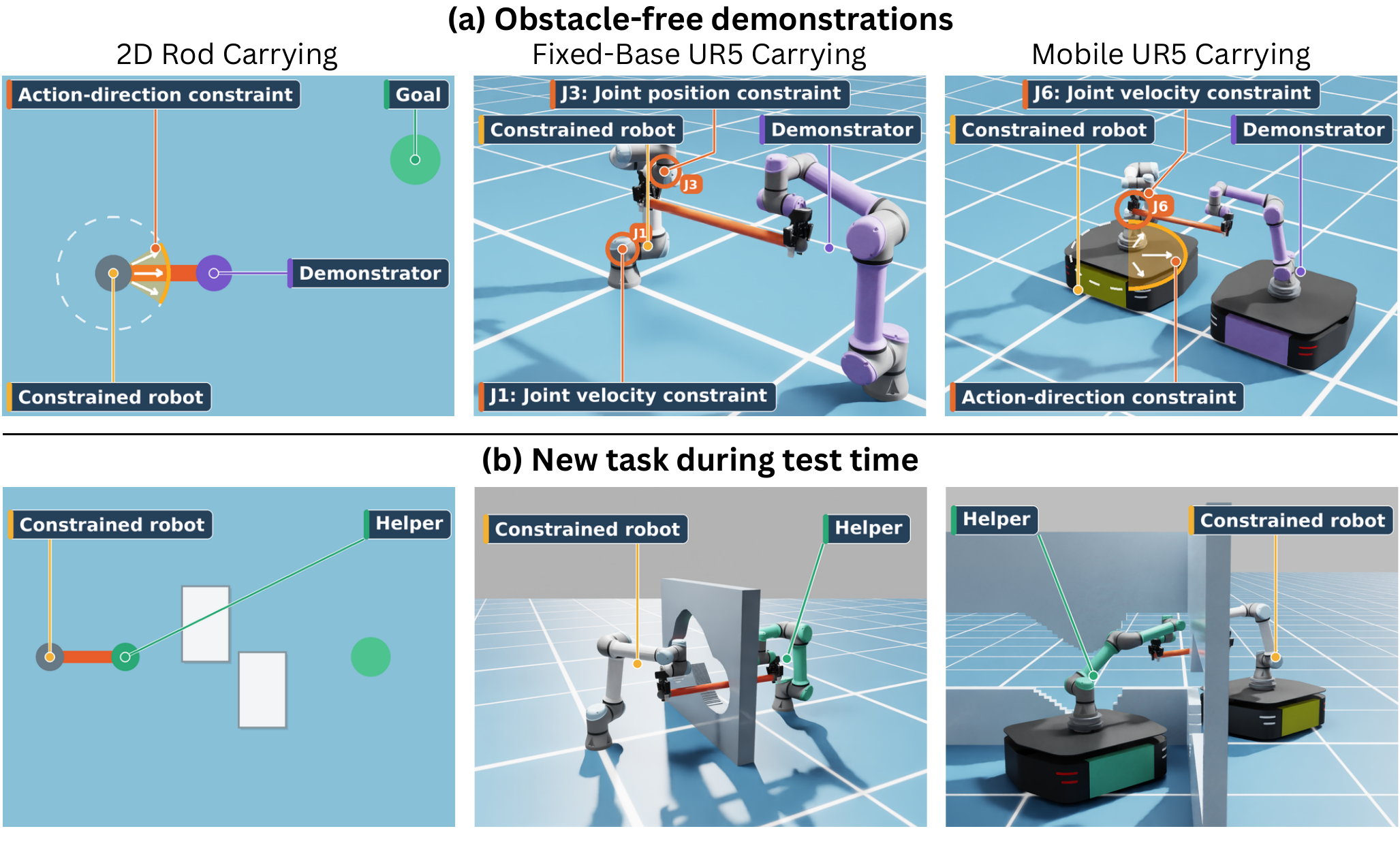}
    \caption{
    \textbf{Overview of our physically coupled partner-adaptation benchmark.}
    (A) During the observation stage, the helper watches a demonstrator coordinate with the constrained agent on an obstacle-free task. The labels show physical constraints on the action space of the constrained agent, which are dependent on the setting. (B) At test time, the helper replaces the demonstrator and coordinates zero-shot with the same constrained agent on a new task with obstacles. The benchmark includes three settings of increasing physical complexity: 2D rod carrying, fixed-base dual-UR5 carrying, and mobile dual-UR5 carrying. Across all settings, the constrained agent's physical constraints remain fixed while the task and helper change.
    }
    \label{fig:tasks}
\end{figure*}

\subsection{Problem Setup}

We consider a helper that must coordinate with a robot whose low-level physical capabilities are initially unknown. We refer to this robot as the \emph{constrained agent}. Its physical actions are restricted by constraints, such as joint position limits, velocity limits, or allowable motion directions. During the observation stage, the helper watches the constrained agent coordinate with another robot, which we call the \emph{demonstrator}. The helper knows the goal of each demonstration and observes the states and actions of both agents, but not the constrained agent's physical constraints. At test time, the helper replaces the demonstrator and must coordinate with the same constrained agent on a new task without first observing how it behaves on that task. The constraints remain fixed while the task and required coordinated motions change, requiring the helper to infer a transferable model of the constrained agent's physical capability from prior coordination.

\subsection{Why Coordinated Behavior is Informative}

A demonstration shows what the constrained agent did, but not what it could have done. If it does not take a particular action, that action may be physically infeasible or simply unnecessary for the task. Inferring its constraints therefore requires reasoning about how the constraints affect the observed behavior. In physically coupled tasks, the demonstrator can help resolve this ambiguity. A constraint on the constrained agent may require the demonstrator to compensate so that the object they are jointly carrying still follows the intended trajectory. This compensation can make the constraint difficult to identify from the constrained agent's actions alone, while the demonstrator's actions provide information about the underlying limitation. Our key insight is that the constrained agent's physical capability shapes the joint behavior of both robots, so both agents' actions provide information for capability inference.

\subsection{Watch, Infer, Coordinate Benchmark}

To evaluate this problem, we construct a benchmark with three joint-carrying settings of increasing physical complexity: 2D rod carrying, fixed-base dual-UR5 carrying, and mobile dual-UR5 carrying. In each setting, the two agents must jointly transport an object from an initial position to a goal. As shown in Figure~\ref{fig:tasks}, demonstrations are collected on obstacle-free tasks requiring different coordinated motions. At test time, obstacles block the direct path to the goal, requiring motions that may not have appeared in the demonstrations. For each setting, we define a fixed set of demonstration tasks and use the same tasks regardless of the constrained agent's physical constraints. The tasks require a range of coordinated motions so that different constraints can affect the observed behavior.

Because physical capability cannot be inferred from arbitrary behavior, we assume a model of how the agents choose their actions. We model the demonstrator and constrained agent as boundedly rational agents coordinating toward a shared task objective \citep{mckelvey1995quantal,lindig2021bounded}. Formally, let \(D\) denote the demonstrator, \(P\) the constrained agent, and \(c\) its capability. At state \(s\), the agents choose actions \(a^D\) and \(a^P\). Each setting defines an action-pair score \(Q_c(s,a^D,a^P)\), computed using the known transition model to score how good the two agents’ actions are for completing the carrying task from state \(s\).

We model the agents using the regularized shared objective
\begin{equation}
\mathcal{J}_c(\pi_D,\pi_P;s)
=
\mathbb{E}_{\substack{a^D\sim\pi_D\\a^P\sim\pi_P}}
\left[
Q_c(s,a^D,a^P)
\right]
-
\tau_D D_{\mathrm{KL}}(\pi_D\|u_D)
-
\tau_P D_{\mathrm{KL}}(\pi_P\|u_P),
\label{eq:shared_team_objective}
\end{equation}
where $D_{\mathrm{KL}}$ denotes the Kullback-Leibler divergence \citep{kullback1951information}, $u_D$ and $u_P$ are uniform over each agent's available actions, and $\tau_D$ and $\tau_P$ control how strongly the agents favor higher-valued actions. This objective gives the mutual soft responses
\begin{equation}
\resizebox{0.98\linewidth}{!}{$\displaystyle
\pi_D^\star(a^D\mid s,c)
\propto
u_D(a^D)
\exp\left(
\frac{
\mathbb{E}_{a^P\sim\pi_P^\star}[Q_c(s,a^D,a^P)]
}{
\tau_D
}
\right),
\qquad
\pi_P^\star(a^P\mid s,c)
\propto
u_P(a^P)
\exp\left(
\frac{
\mathbb{E}_{a^D\sim\pi_D^\star}[Q_c(s,a^D,a^P)]
}{
\tau_P
}
\right)
$}
\label{eq:mutual_soft_responses}
\end{equation}
The two policies depend on one another. Restricting the constrained agent's feasible actions changes both its own behavior and how the demonstrator is expected to respond, capturing the compensatory behavior described above. Together with the physical feasibility rules of each setting, these policies define a distribution \(\rho_c(a^D,a^P\mid s)\) over action pairs, which we use to generate demonstrations under the true capability.

At test time, the helper replaces the demonstrator while the constrained agent retains the same capability and behavioral model. The agents choose their actions simultaneously, and the resulting action pair is executed according to the dynamics and feasibility rules of the setting. We provide details on computing the mutual responses in Appendix~\ref{app:behavioral_model}.

The benchmark spans three joint-carrying settings of increasing physical complexity. We begin with 2D rod carrying, where two agents apply forces at opposite ends of a rigid rod and jointly determine its translation and rotation. The constrained agent is limited to a continuous range of force directions, and we consider 2,665 candidate force-direction constraints. We then move to fixed-base dual-UR5 carrying, where two robot arms grasp opposite ends of a rigid beam and the constrained agent is subject to joint position and joint velocity limits. Because the joint motion required to move the beam depends on the robot's posture, the same beam motion may be feasible in one configuration and infeasible in another. This setting contains 172 candidate capabilities over individual and paired joint limitations. Finally, mobile dual-UR5 carrying adds mobile bases to the two arms, introducing constraints on both arm and base motion. Combining the possible arm and base constraints gives 475 candidate capabilities. In each setting, we use three fixed physical-constraint cases for final evaluation. Additional benchmark specifications and evaluation details are provided in Appendix~\ref{app:experiment_details}.

\section{Method}
\label{sec:method}

\subsection{Overview}

We infer the constrained agent's capability by treating each candidate capability as a hypothesis about which actions the agent can execute. At each observed state, a candidate changes the constrained agent's feasible actions and, in turn, how the demonstrator is expected to respond. Using the boundedly rational behavioral model, we predict the joint behavior of the two agents under each candidate and score how well it matches the observed actions across demonstrations. These scores are used to form a belief over the constrained agent's capability. In the low-dimensional setting, we can compute this score explicitly. In the more complex settings, evaluating the behavioral model for every observed state and candidate becomes too expensive, so we use learned models to estimate how well each candidate explains the observed coordination.

Formally, let $\mathcal{D}$ denote the observed demonstrations and $\mathcal{C}$ the set of candidate capabilities. We define $S(c,\mathcal{D})$ as the score measuring how well candidate $c$ explains the observed coordination across all demonstrations. Let $p(c)$ denote the prior probability assigned to candidate capability $c$ before observing the demonstrations. We combine this prior with the demonstration score to obtain
\begin{equation}
L(c)
=
\log p(c) + S(c,\mathcal{D}),
\qquad
b(c \mid \mathcal{D})
=
\operatorname{softmax}_{c \in \mathcal{C}} L(c),
\label{eq:capability_belief_overview}
\end{equation}
where $b(c \mid \mathcal{D})$ is the resulting belief over the constrained agent's capability. Algorithm~\ref{alg:ours} summarizes the overall procedure.

\begin{algorithm}[t]
\caption{\ours{} overview}
\label{alg:ours}
\begin{algorithmic}[1]

\Statex \textbf{Input:} demonstrations $\mathcal{D}$, candidate capabilities $\mathcal{C}$, prior $p(c)$, test task

\For{each capability $c \in \mathcal{C}$}
    \State Compute $S(c,\mathcal{D})$ from the observed actions of both agents
    \State $L(c) \gets \log p(c) + S(c,\mathcal{D})$
\EndFor

\State $b(c \mid \mathcal{D}) \gets
\operatorname{softmax}_{c \in \mathcal{C}} L(c)$
\State Plan with the constrained agent using $b(c \mid \mathcal{D})$

\end{algorithmic}
\end{algorithm}

To compute $S(c,\mathcal{D})$, we measure how likely the observed actions of both agents are under candidate capability $c$. At each observed state, the behavioral model defines a distribution $\rho_{c,\tau}(a^D,a^P \mid s)$ over their joint actions, where $\tau=(\tau_D,\tau_P)$ contains the agents' behavior temperatures. We evaluate the probability assigned to each observed action pair and combine these probabilities across all states and demonstrations:
\begingroup\small
\begin{equation}
S(c,\mathcal{D})
=
\log \left(
\sum_{\tau}
\exp \left[
\sum_{d \in \mathcal{D}}
\sum_{t \in d}
\log \rho_{c,\tau}
\left(
a_t^D,a_t^P \mid s_t
\right)
+
\log p(\tau)
\right]
\right).
\label{eq:capability_score}
\end{equation}
\endgroup
A candidate therefore receives a high score when it predicts both the constrained agent's actions and the demonstrator's response well.

\subsection{Computing Capability Scores}

Computing the score in Eq.~\ref{eq:capability_score} exactly requires solving the behavioral model at every observed state for every candidate capability. This becomes prohibitively expensive as the state and action spaces grow. We therefore train learned models offline on simulated coordination where the constrained agent's capability is known. At inference time, the true capability is hidden, and the models receive the observed coordination together with a candidate capability to estimate how well that candidate explains the demonstrations.

Across all three settings, we use the same capability-scoring idea, but vary how much of the computation is approximated by a learned model. In the 2D rod setting, we learn the action values predicted by the behavioral model for both agents, convert them into action distributions, and compute the capability score explicitly from the likelihood of the observed actions. In the fixed-base dual-UR5 setting, we predict each observed step's contribution to the same capability score directly and combine these predictions across the demonstration. The mobile dual-UR5 setting further increases the complexity and introduces both arm and base constraints, so we predict scores at the demonstration level and combine the contributions of the arm and base components when evaluating a candidate capability. The resulting scores are combined into the capability belief used for test-time planning.

\subsection{Planning with the Inferred Capability}

At test time, the helper must coordinate with the constrained agent on a new task that may require motions not seen in the demonstrations. Rather than transferring a demonstrated coordination strategy, we use the inferred capability to constrain the partner behavior considered during planning. Within a model predictive control (MPC) framework, we use the cross-entropy method (CEM) to optimize the helper's action sequence \citep{rubinstein1999cross,rubinstein2004cross}. For each candidate helper sequence, the planner rolls out how the constrained agent is expected to respond under its inferred capability. During each rollout, the constrained agent is limited to actions that are feasible under the inferred capability. The planner therefore searches over coordination strategies that are compatible with what the constrained agent can physically execute, allowing the helper to plan for a new task without observing prior coordination on that task. We provide additional derivations and implementation details in Appendix~\ref{app:method_details}.

\section{Experiments and Results}
\label{sec:experiments}

We evaluate whether our method can accurately infer the constrained agent's physical constraints from prior coordination and whether the inferred capability supports zero-shot coordination on a new task. We study both questions across the 2D rod, fixed-base dual-UR5, and mobile dual-UR5 settings.

\subsection{Experimental Setup}
For capability inference, we vary the number of observed demonstrations over $N \in \{1,2,4,8\}$ and measure normalized Hamming distance between the inferred and true constraints, where lower is better. For zero-shot coordination, all inference-based methods use $N=8$ demonstrations. We evaluate performance on the new tasks with obstacles and additionally report results on obstacle-free control tasks to separate failures in basic coordination from failures that arise when the new task requires different motions.

We compare against four baselines that isolate different sources of performance. CE-CM-Div \citep{tisnikar2026partner} is a prior capability-inference method that estimates task-invariant partner capabilities from multi-agent trajectories and serves as our primary inference baseline. Capacity-blind CEM uses the same planner as our method but receives no information about the constrained agent's capability, testing whether capability information is necessary for planning. Oracle CEM receives the true physical constraints and provides a reference for performance with perfect capability information. Behavioral cloning directly learns from the observed demonstrations, allowing us to test whether reproducing demonstrated coordination is sufficient when the task changes.

For planning, we report task success and Success weighted by Number of Actions (SNA) \citep{chen2020learning}. Failed episodes receive an SNA of zero. For successful episodes, SNA is the ratio between the reference number of actions and the larger of the executed and reference numbers of actions, so higher SNA indicates more efficient successful coordination. Success and SNA are reported as percentages, with higher values indicating better performance. All results are reported as mean and standard error over three random seeds. All CEM-based methods use the same planner and differ only in the capability information provided to it. Additional details on baseline implementations, evaluation tasks, training, and planner hyperparameters are provided in Appendix~\ref{app:experiment_details}.

\subsection{Constraint Inference}

We first measure constraint-inference accuracy from prior coordination across all three settings and demonstration budgets. Table~\ref{tab:constraint_inference} shows that our method outperforms CE-CM-Div across all three settings and demonstration budgets. Additional demonstrations are especially helpful in the mobile dual-UR5 setting, where error decreases from $0.190$ at $N=1$ to $0.009$ at $N=8$.

We also perform an ablation to test whether observing the demonstrator improves capability inference. For this ablation, we retrain the inference model using only the constrained agent's behavior. As shown in Table~\ref{tab:constraint_inference}, using both agents' behavior roughly halves inference error in the fixed-base dual-UR5 setting across all demonstration budgets and reduces error in the mobile dual-UR5 setting at every budget. In the 2D rod setting, the demonstrator's behavior substantially improves inference when only one or two demonstrations are available. These results support our key insight that physical coupling makes the demonstrator's response informative about constraints that may be difficult to identify from the constrained agent's behavior alone.

\begin{table}[h]
    \centering
    \caption{
        Constraint-inference accuracy at $N\in\{1,2,4,8\}$ demonstrations.
        We report normalized Hamming distance between the inferred and
        true constraints.
        ``Ours, partner only'' uses only the constrained agent's behavior
        for inference and excludes the demonstrator's behavior.
        Values are mean $\pm$ standard error; lower is better.
        Bold indicates the best result within each robot setting and row,
        including ties.
    }
    \label{tab:constraint_inference}
    \setlength{\tabcolsep}{3pt}
    \renewcommand{\arraystretch}{1.1}

    \resizebox{\linewidth}{!}{%
    \begin{tabular}{c ccc ccc ccc}
        \toprule
        & \multicolumn{3}{c}{2D Rod}
        & \multicolumn{3}{c}{Fixed-Base Dual-UR5}
        & \multicolumn{3}{c}{Mobile Dual-UR5} \\
        \cmidrule(lr){2-4}
        \cmidrule(lr){5-7}
        \cmidrule(lr){8-10}

        $N$
        & CE-CM-Div
        & \shortstack{Ours,\\partner only}
        & Ours
        & CE-CM-Div
        & \shortstack{Ours,\\partner only}
        & Ours
        & CE-CM-Div
        & \shortstack{Ours,\\partner only}
        & Ours \\
        \midrule

        1
        & $0.304 \pm 0.038$
        & $0.104 \pm 0.015$
        & \textbf{0.058 $\pm$ 0.017}
        & $0.074 \pm 0.000$
        & $0.068 \pm 0.006$
        & \textbf{0.031 $\pm$ 0.012}
        & $0.306 \pm 0.028$
        & $0.201 \pm 0.045$
        & \textbf{0.190 $\pm$ 0.060} \\

        2
        & $0.340 \pm 0.041$
        & $0.031 \pm 0.009$
        & \textbf{0.003 $\pm$ 0.002}
        & $0.074 \pm 0.000$
        & $0.037 \pm 0.000$
        & \textbf{0.019 $\pm$ 0.000}
        & $0.278 \pm 0.028$
        & $0.148 \pm 0.042$
        & \textbf{0.130 $\pm$ 0.060} \\

        4
        & $0.348 \pm 0.031$
        & \textbf{0.002 $\pm$ 0.002}
        & \textbf{0.002 $\pm$ 0.002}
        & $0.074 \pm 0.000$
        & $0.037 \pm 0.000$
        & \textbf{0.019 $\pm$ 0.000}
        & $0.250 \pm 0.000$
        & $0.090 \pm 0.006$
        & \textbf{0.069 $\pm$ 0.005} \\

        8
        & $0.302 \pm 0.011$
        & \textbf{0.000 $\pm$ 0.000}
        & $0.002 \pm 0.002$
        & $0.074 \pm 0.000$
        & $0.043 \pm 0.006$
        & \textbf{0.019 $\pm$ 0.011}
        & $0.250 \pm 0.000$
        & $0.042 \pm 0.000$
        & \textbf{0.009 $\pm$ 0.000} \\

        \bottomrule
    \end{tabular}%
    }
\end{table}

\subsection{Zero-Shot Coordination}
We next evaluate whether the inferred capability supports coordination on the new tasks with obstacles. As shown in Table~\ref{tab:planning_results}, our method achieves success rates of $92.6\%$, $85.2\%$, and $63.0\%$ in the 2D rod, fixed-base dual-UR5, and mobile dual-UR5 settings, respectively, approaching Oracle CEM at $96.3\%$, $88.9\%$, and $70.4\%$. CE-CM-Div performs substantially worse across all three settings, showing that more accurate capability inference translates into better coordination when the task changes.

The remaining baselines highlight why capability information must transfer beyond the demonstrated behavior. Capacity-blind CEM performs worse when successful planning depends on accounting for the constrained agent's limitations, while behavioral cloning fails on all three new tasks despite nonzero performance on the obstacle-free controls. On the obstacle-free control tasks, our method closely matches Oracle CEM across all three settings.

\begin{table*}[h]
    \centering
    \caption{
        Planning performance on the obstacle-free control and new tasks
        with obstacles. Our method, CE-CM-Div, and behavioral cloning use
        \(N=8\) demonstrations. Capacity-blind CEM uses no demonstrations,
        while Oracle CEM receives the true constraints.
        Success and SNA are percentages reported as mean \(\pm\)
        standard error over three seeds; higher is better.
        Bold indicates the best non-oracle result for each task and metric,
        including ties.
    }
    \label{tab:planning_results}
    \setlength{\tabcolsep}{4pt}
    \resizebox{\textwidth}{!}{%
    \begin{tabular}{llcccccc}
        \toprule
        & &
        \multicolumn{2}{c}{2D Rod}
        & \multicolumn{2}{c}{Fixed-Base Dual-UR5}
        & \multicolumn{2}{c}{Mobile Dual-UR5} \\
        \cmidrule(lr){3-4}
        \cmidrule(lr){5-6}
        \cmidrule(lr){7-8}
        Task & Method
        & Success $\uparrow$ & SNA $\uparrow$
        & Success $\uparrow$ & SNA $\uparrow$
        & Success $\uparrow$ & SNA $\uparrow$ \\
        \midrule

        \multirow{5}{*}{\shortstack{Obstacle-free\\control}}
        & Capacity-blind + CEM
        & \textbf{100.0 $\pm$ 0.0}
        & $91.4 \pm 0.9$
        & $0.0 \pm 0.0$
        & $0.0 \pm 0.0$
        & $66.7 \pm 0.0$
        & $55.2 \pm 0.0$ \\

        & CE-CM-Div + CEM
        & \textbf{100.0 $\pm$ 0.0}
        & $90.4 \pm 1.6$
        & $55.6 \pm 11.1$
        & $30.5 \pm 8.3$
        & $66.7 \pm 0.0$
        & $54.3 \pm 0.0$ \\

        & Behavioral cloning
        & $66.7 \pm 0.0$
        & $52.5 \pm 6.1$
        & $22.2 \pm 11.1$
        & $7.5 \pm 3.8$
        & $66.7 \pm 0.0$
        & $31.1 \pm 0.0$ \\

        & \fade{Oracle + CEM}
        & \fade{$100.0 \pm 0.0$}
        & \fade{$91.4 \pm 0.2$}
        & \fade{$100.0 \pm 0.0$}
        & \fade{$44.8 \pm 1.8$}
        & \fade{$100.0 \pm 0.0$}
        & \fade{$80.2 \pm 0.0$} \\
        \cmidrule(lr){2-8}

        & \textbf{Ours + CEM}
        & \textbf{100.0 $\pm$ 0.0}
        & \textbf{92.4 $\pm$ 0.4}
        & \textbf{100.0 $\pm$ 0.0}
        & \textbf{43.9 $\pm$ 0.8}
        & \textbf{100.0 $\pm$ 0.0}
        & \textbf{80.2 $\pm$ 0.0} \\

        \midrule

        \multirow{5}{*}{\shortstack{New tasks\\with obstacles}}
        & Capacity-blind + CEM
        & $51.9 \pm 3.7$
        & $44.4 \pm 2.8$
        & $0.0 \pm 0.0$
        & $0.0 \pm 0.0$
        & $25.9 \pm 9.8$
        & $14.2 \pm 5.4$ \\

        & CE-CM-Div + CEM
        & $55.6 \pm 0.0$
        & $46.1 \pm 0.7$
        & $44.4 \pm 0.0$
        & $16.9 \pm 1.5$
        & $37.0 \pm 3.7$
        & $19.7 \pm 1.8$ \\

        & Behavioral cloning
        & $0.0 \pm 0.0$
        & $0.0 \pm 0.0$
        & $0.0 \pm 0.0$
        & $0.0 \pm 0.0$
        & $0.0 \pm 0.0$
        & $0.0 \pm 0.0$ \\

        & \fade{Oracle + CEM}
        & \fade{$96.3 \pm 3.7$}
        & \fade{$76.0 \pm 2.0$}
        & \fade{$88.9 \pm 6.4$}
        & \fade{$31.3 \pm 1.9$}
        & \fade{$70.4 \pm 3.7$}
        & \fade{$38.4 \pm 2.5$} \\
        \cmidrule(lr){2-8}

        & \textbf{Ours + CEM}
        & \textbf{92.6 $\pm$ 3.7}
        & \textbf{72.0 $\pm$ 3.3}
        & \textbf{85.2 $\pm$ 3.7}
        & \textbf{31.1 $\pm$ 0.8}
        & \textbf{63.0 $\pm$ 7.4}
        & \textbf{34.5 $\pm$ 4.4} \\

        \bottomrule
    \end{tabular}%
    }
\end{table*}

\subsection{Qualitative Examples}

\begin{figure*}[h]
    \centering
    \includegraphics[width=\textwidth, trim={0cm 11cm 2cm 0}, clip]{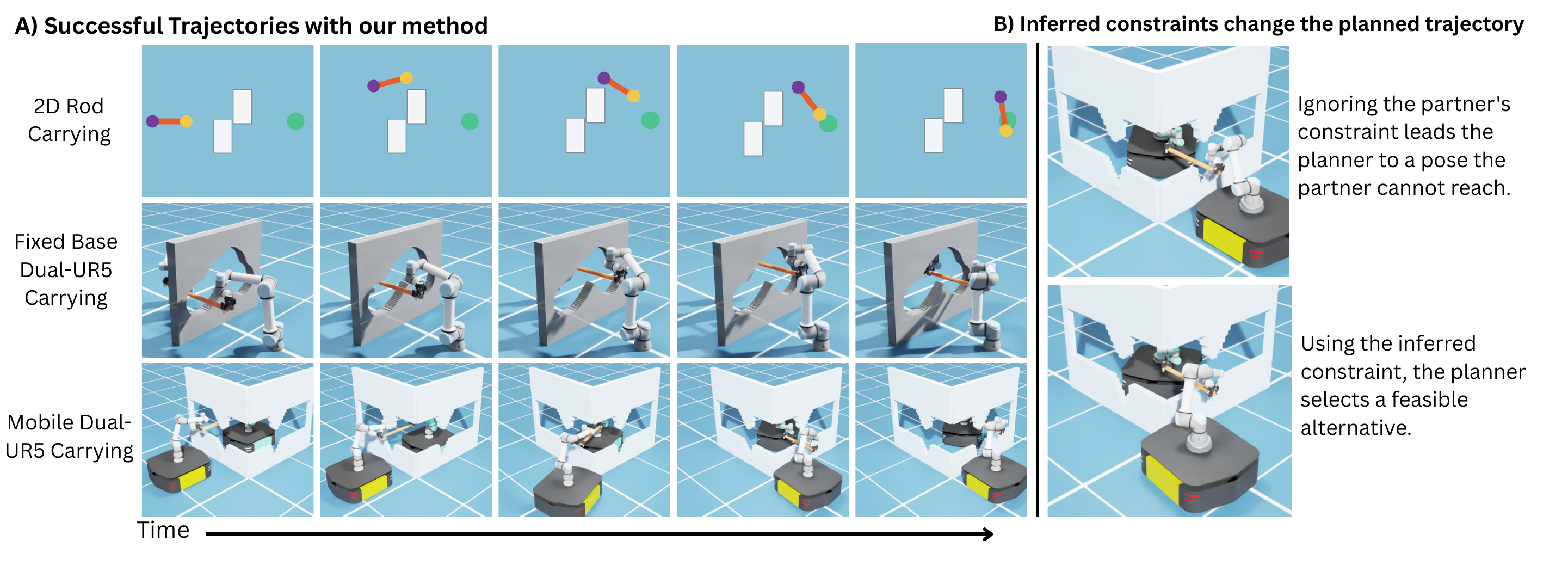}
    \caption{\textbf{Qualitative examples of zero-shot coordination using our method.} (A) Representative successful trajectories for 2D rod carrying, fixed-base dual-UR5 carrying, and mobile dual-UR5 carrying, shown from left to right over time. (B) Example of how an inferred physical constraint changes test-time planning. Ignoring the constrained agent's limitation leads to a pose it cannot reach, while incorporating the inferred constraint causes the planner to select a feasible alternative.
    }
    \label{fig:qualitative}
\end{figure*}

Figure~\ref{fig:qualitative} provides qualitative examples of how the inferred capability affects test-time coordination. Panel A shows successful trajectories across all three benchmark settings, where the helper coordinates with the constrained agent to navigate the jointly carried object around obstacles. Panel B illustrates how the inferred constraint changes the planner's behavior. When the partner's constraint is ignored, the planner selects a motion that requires the constrained agent to reach a natural but infeasible pose. Using the inferred constraint, the planner instead selects a feasible alternative that respects the partner's physical limitations.

\section{Conclusion and Limitations}

In summary, we formulate the problem of inferring a robot partner's low-level physical constraints from prior multi-agent coordination and using them for zero-shot coordination on a new task. We introduce \ours{}, a benchmark spanning three physically coupled carrying settings, and show how the joint behavior of both agents can be used to infer a partner's physical constraints. Our experiments show that the inferred constraints improve zero-shot coordination and achieve performance close to an oracle with access to the true constraints.

As a first step, we intentionally assume a known behavioral model, goals and dynamics, full observations, and a finite constraint set to isolate the core challenge of constraint inference and coordination without introducing additional sources of uncertainty. Relaxing these assumptions is an important next step toward broader generalization. Moreover, directly comparable baselines are limited, as existing approaches use different capability representations or observations. We will release benchmark code and evaluation scripts to support broader evaluations across tasks, constraints, and partner behaviors, and future extensions to heterogeneous teams, including human-robot coordination.

\bibliography{iclr2027_conference}
\bibliographystyle{iclr2027_conference}

\appendix
\section{Appendix}

\subsection{Method Details}
\label{app:method_details}
\subsubsection{Behavioral Model}
\label{app:behavioral_model}

The capability score in Eq.~\ref{eq:capability_score} requires predicting how the two agents would behave under a candidate capability $c$. We now describe the behavioral model used to construct this prediction. Intuitively, a candidate capability determines which actions the constrained agent can take. Given these feasible actions, each agent favors actions that lead to good joint outcomes, without assuming perfectly optimal action selection. The agents' predicted actions are interdependent: the demonstrator chooses actions based on what the constrained agent can do, while the constrained agent's preferred actions depend on how the demonstrator is expected to respond. We therefore update their predicted policies iteratively to obtain a model of their joint behavior.

Let $D$ denote the demonstrator and $P$ the constrained agent. At state $s$, the agents choose actions $a^D$ and $a^P$. Each setting defines a shared action-pair value
\[
Q_c(s,a^D,a^P),
\]
which measures the value of taking the joint action $(a^D,a^P)$ and continuing from the resulting state. The candidate capability $c$ determines the feasible action set $\mathcal{A}_P(s,c)$ for the constrained agent. Thus, different candidate capabilities can produce different predicted behavior even at the same state.

We model both agents as boundedly rational under the shared regularized objective
\begin{equation}
\mathcal{J}_c(\pi_D,\pi_P;s)
=
\mathbb{E}_{a^D\sim\pi_D,\;a^P\sim\pi_P}
\left[
Q_c(s,a^D,a^P)
\right]
-
\tau_D D_{\mathrm{KL}}(\pi_D\|u_D)
-
\tau_P D_{\mathrm{KL}}(\pi_P\|u_{P,c}),
\label{eq:behavioral_objective}
\end{equation}
where $u_D$ is the uniform distribution over the demonstrator's available actions and $u_{P,c}$ is the uniform distribution over the constrained agent's actions that are feasible under capability $c$. The temperatures $\tau_D$ and $\tau_P$ determine how strongly the agents favor high-value actions. Smaller temperatures place more probability on the highest-value actions, while larger temperatures allow a wider range of actions.

The two agents' policies are coupled because whether an action is useful for one agent depends on what the other agent is expected to do. We therefore compute the expected value of each action by averaging over the other agent's current policy:
\begin{equation}
\bar{Q}_D(s,a^D)
=
\mathbb{E}_{a^P\sim\pi_P}
\left[
Q_c(s,a^D,a^P)
\right],
\qquad
\bar{Q}_P(s,a^P)
=
\mathbb{E}_{a^D\sim\pi_D}
\left[
Q_c(s,a^D,a^P)
\right].
\label{eq:averaged_action_values}
\end{equation}
For example, $\bar{Q}_D(s,a^D)$ measures how good demonstrator action $a^D$ is on average given what we currently expect the constrained agent to do.

We convert these expected action values into soft action distributions:
\begin{equation}
\widetilde{\pi}_D(a^D)
=
\frac{
u_D(a^D)\exp\left(\bar{Q}_D(s,a^D)/\tau_D\right)
}{
\sum_{a^{D\prime}}
u_D(a^{D\prime})
\exp\left(\bar{Q}_D(s,a^{D\prime})/\tau_D\right)
},
\label{eq:demonstrator_soft_response}
\end{equation}
and
\begin{equation}
\widetilde{\pi}_P(a^P)
=
\frac{
u_{P,c}(a^P)\exp\left(\bar{Q}_P(s,a^P)/\tau_P\right)
}{
\sum_{a^{P\prime}}
u_{P,c}(a^{P\prime})
\exp\left(\bar{Q}_P(s,a^{P\prime})/\tau_P\right)
}.
\label{eq:partner_soft_response}
\end{equation}
These distributions assign greater probability to actions with higher expected joint value. Because $u_{P,c}$ places probability only on actions that are feasible under capability $c$, infeasible constrained-agent actions receive no probability.

The two soft responses depend on one another. Changing the constrained agent's policy changes the demonstrator's expected action values, which can change the demonstrator's policy, and vice versa. We therefore compute the policies iteratively. Starting from uniform policies, we first update the demonstrator based on the constrained agent's current policy and then update the constrained agent based on the newly updated demonstrator policy. To make these updates more stable, we move each policy partway toward its newly computed soft response. For the demonstrator, this update is
\begin{equation}
\log \pi_D^{(m)}
=
\operatorname{lognormalize}
\left(
(1-\eta)\log \pi_D^{(m-1)}
+
\eta\log\widetilde{\pi}_D^{(m)}
\right),
\label{eq:demonstrator_policy_update}
\end{equation}
and the constrained agent is updated analogously:
\begin{equation}
\log \pi_P^{(m)}
=
\operatorname{lognormalize}
\left(
(1-\eta)\log \pi_P^{(m-1)}
+
\eta\log\widetilde{\pi}_P^{(m)}
\right),
\label{eq:partner_policy_update}
\end{equation}
where $\eta=0.5$ and $\operatorname{lognormalize}(x)=x-\operatorname{logsumexp}(x)$. Algorithm~\ref{alg:mutual_response} summarizes the complete procedure.

\begin{algorithm}[h]
\caption{Mutual soft-response computation}
\label{alg:mutual_response}
\begin{algorithmic}[1]

\Statex \textbf{Input:} state $s$, candidate capability $c$, joint-action values $Q_c$, temperatures $\tau_D,\tau_P$
\State Initialize $\pi_D^{(0)} \gets u_D$ and $\pi_P^{(0)} \gets u_{P,c}$

\For{$m=1,\ldots,M$}
    \State Compute $\bar{Q}_D^{(m)}$ using $\pi_P^{(m-1)}$ and Eq.~\ref{eq:averaged_action_values}
    \State Compute $\widetilde{\pi}_D^{(m)}$ using Eq.~\ref{eq:demonstrator_soft_response}
    \State Update $\pi_D^{(m)}$ using Eq.~\ref{eq:demonstrator_policy_update}

    \State Compute $\bar{Q}_P^{(m)}$ using $\pi_D^{(m)}$ and Eq.~\ref{eq:averaged_action_values}
    \State Compute $\widetilde{\pi}_P^{(m)}$ using Eq.~\ref{eq:partner_soft_response}
    \State Update $\pi_P^{(m)}$ using Eq.~\ref{eq:partner_policy_update}
\EndFor

\State Compute a final soft response $\pi_D$ using $\pi_P^{(M)}$
\State Compute a final soft response $\pi_P$ using $\pi_D^{(M)}$

\Statex \textbf{Return:} $\pi_D,\pi_P$

\end{algorithmic}
\end{algorithm}

After the iterative updates, we compute one final soft response for each agent using the final policy of the other agent. We denote these resulting policies by $\pi_D$ and $\pi_P$. Together, they define the capability-conditioned joint-action distribution
\begin{equation}
\rho_{c,\tau}(a^D,a^P\mid s)
=
\pi_D(a^D\mid s,c,\tau)
\,
\pi_P(a^P\mid s,c,\tau),
\label{eq:behavioral_joint_distribution}
\end{equation}
where $\tau=(\tau_D,\tau_P)$. Changing the candidate capability $c$ therefore changes the constrained agent's feasible actions and, through the coupled policies, can also change the demonstrator's predicted behavior.

\subsubsection{Capability Inference Details}

Across all three settings, capability inference follows the same basic procedure. For each candidate capability, we compute a score measuring how well that candidate explains the observed coordination. Candidates that are physically inconsistent with the demonstrations are excluded, and the remaining scores are combined with a capability prior and normalized to form the belief used for planning. The settings differ only in how we compute the score.

In the 2D rod setting, the action space is small enough to score candidates from the predicted actions directly. For each observed state and candidate capability, the learned critic predicts values for the 64 possible force directions of both the demonstrator and constrained agent. We convert these values into action distributions and evaluate how much probability they assign to the two actions that were actually observed. Because the recorded actions are continuous directions, we interpret each discrete action probability as a density over its corresponding angular interval. We additionally mix the density with a small uniform component of weight $0.02$ so that a single unusual action does not assign zero likelihood to an otherwise plausible candidate.

We sum the log probabilities of both agents' observed actions over all steps and demonstrations to obtain a score for each candidate. The behavioral temperatures of the two agents are not assumed to be known. We therefore evaluate seven possible temperatures for each agent,
\[
\{0.001,\;0.002,\;0.005,\;0.01,\;0.02,\;0.05,\;0.1\},
\]
giving 49 temperature pairs. A single temperature pair is shared across all selected demonstrations, and we marginalize over these 49 possibilities when computing the final candidate score. The resulting scores are combined with the capability prior and normalized across candidates.

For the UR5 settings, explicitly constructing these action distributions at every observed state for every candidate would require substantially more computation. We therefore learn the candidate score directly. In the fixed-base setting, a candidate-conditioned critic assigns a scalar score to each observed step using the state, the actions of both agents, their resulting motion, and the candidate capability. We sum these step scores over each demonstration and then combine them across all selected demonstrations. A separate physical-consistency check removes candidates whose joint position or velocity limits are incompatible with the observed motion. We then add a sparsity prior that favors capabilities with fewer active joint constraints and normalize the resulting scores over the 172 candidates.

The mobile setting uses the same candidate-scoring idea, but scores complete observed sequences rather than individual steps. A recurrent critic processes each demonstration and produces a score from the resulting sequence representation. Because a mobile capability contains both an arm constraint and a base-motion constraint, the critic scores these two components separately and then combines their contributions when evaluating a complete candidate. As in the fixed-base setting, an independent physical-consistency check removes candidates that cannot reproduce the observed motion before the learned scores are normalized. This produces a distribution over the 475 composite capabilities.

\subsubsection{Physical Consistency and Test-Time Planning}

At test time, the inferred capability is used inside CEM-MPC to predict which coordinated motions are physically possible for the constrained agent. For each candidate helper action sequence, the planner predicts the constrained agent's response while restricting it to actions that are feasible under the capability being considered. The resulting joint motion is rolled forward and scored for progress toward the test-task goal. In the rod and fixed-base settings, planning accounts for uncertainty over multiple candidate capabilities from the inferred belief, while the mobile setting plans with the highest-scoring capability. The helper executes the first action of the selected sequence and replans from the resulting observed state.

Proposed motions must also satisfy the physical constraints of the setting. In the UR5 settings, the planner checks whether a motion is compatible with the candidate capability as well as the robot kinematics, workspace, and collision constraints. In the rod setting, proposed motions are similarly checked for collisions. Motions that violate these constraints are treated as infeasible rather than modified into a different executable motion.

Within each setting, all CEM-based methods use the same planner and planning budget. They differ only in the capability information supplied to the planner: our method uses its inferred capability, CE-CM-Div uses its inferred capability, Capacity-blind CEM receives no demonstration-based capability information, and Oracle CEM receives the true capability.

\subsection{Experiment Implementation Details}
\label{app:experiment_details}

Across all three settings, we evaluate capability inference using $N\in\{1,2,4,8\}$ obstacle-free demonstrations and use $N=8$ demonstrations for zero-shot coordination. Each setting contains three fixed physical-constraint cases for final evaluation. Planning is evaluated on three new tasks with obstacles and one obstacle-free control task, using three evaluation seeds.

\subsubsection{2D Rod Carrying}

In the 2D rod setting, the state consists of the rod center and orientation, $(x,y,\theta)$. The two agents apply unit forces at opposite ends of the rod, jointly determining its translation and rotation. Each agent acts over 64 force directions. The demonstrator can apply force in any direction, while the constrained agent is limited to a continuous angular range specified by its capability. The candidate set contains 2,665 such constraints, and the three evaluation cases use angular ranges centered at $0^\circ$ with widths of $30^\circ$, $45^\circ$, and $65^\circ$. Demonstrations are generated on obstacle-free goal-reaching tasks using the behavioral model described above. For $N=1,2,4,8$, we use increasingly larger subsets of the same fixed demonstration sequence, so additional demonstrations are added without changing the earlier ones. To make inference efficient, we train a neural network to predict the action values produced by the behavioral model for both agents. The model uses three hidden layers of width 256 and is trained on 98,304 simulated examples pairing rod states and candidate capabilities with action-value targets from the behavioral model.

\subsubsection{Fixed-Base Dual-UR5 Carrying}

The fixed-base dual-UR5 setting is simulated in Isaac Sim with two UR5e arms grasping opposite ends of a rigid beam. Demonstrations represent the coordinated end-effector motion using the beam's $y$ and $z$ position and yaw orientation. Each agent selects from 257 possible three-dimensional beam-velocity commands, and the two proposals are combined into a shared beam motion. A capability describes possible restrictions on the six robot joints: each joint can have a reduced velocity limit, a tighter lower position limit, or a tighter upper position limit. This gives 18 possible individual restrictions. The 172 candidate capabilities include the unrestricted robot, each of the 18 individual restrictions, and every pair of restrictions. The three evaluation cases use a velocity restriction on the first joint, a lower position restriction on the third joint, and the combination of those two restrictions. The obstacle-free demonstrations cover six different beam motions, with $N=8$ using eight demonstrations drawn from these motions. Capability scoring uses a neural network conditioned on both the observed coordination and a candidate capability. It produces a score for each observed step, and these scores are combined across the demonstrations. The model is trained with AdamW for 12,000 updates using demonstrations generated under known capabilities.

\subsubsection{Mobile Dual-UR5 Carrying}

The mobile dual-UR5 setting extends the same manipulation task with planar mobile bases. The state has 22 dimensions and contains the beam pose, both base poses, and both six-joint arm configurations. Each agent selects a four-dimensional motion command controlling translation and yaw from a set of 1,281 possible actions. A capability combines a constraint on the directions in which the base can move with a constraint on arm motion. We consider 25 base constraints and 19 arm constraints, producing 475 possible combinations. Each evaluation case is paired with eight fixed obstacle-free demonstrations totaling 100 observed steps. Six demonstrations emphasize arm motion while the bases remain stationary, and two emphasize base motion while the arms are held fixed. For $N=1,2,4$, we use fixed subsets of these eight demonstrations. Capability scoring uses a recurrent neural network that processes an entire demonstration at once. It scores the arm and base components of each candidate and combines them to obtain a score for the complete capability. The model uses a one-layer GRU with hidden width 48 and is trained on 25 capability cases, with three additional cases used for validation.

\subsubsection{Baseline Implementations}

CE-CM-Div is our primary capability-inference baseline. It receives the same obstacle-free demonstrations as our method but infers capability from the observed state trajectories rather than the actions of both agents. For each candidate capability, it generates possible trajectories and compares them with the observed trajectory, then projects the resulting estimate onto the same candidate set used by our method. The inferred capability is passed to the same setting-specific CEM-MPC planner used by our method. In the mobile setting, the arm and base components are inferred separately and then combined into a complete capability.

Capacity-blind CEM uses the same planner but receives no information from the demonstrations about the constrained agent's capability. It therefore plans from a fixed, task-independent capability assumption defined for each setting. This corresponds to an unrestricted force-direction model in the rod setting, the fixed capability prior in the fixed-base setting, and the nominal composite capability in the mobile setting. Oracle CEM instead receives the true capability while using the same planner and planning budget, providing a reference for performance with perfect capability information.

Behavioral cloning directly predicts the helper's action from the $N=8$ obstacle-free demonstrations. For each setting, we train a three-layer width-256 policy to predict the helper action in that setting's action space. At test time, the policy acts directly without capability inference or CEM-MPC planning.

For the partner-only comparison, we train separate inference models using only the constrained agent's behavior. These models use the same capability candidates and evaluation demonstrations as the full method but exclude the demonstrator's behavior when inferring the constraint.

\subsubsection{Evaluation Protocol}

Capability inference is evaluated using normalized Hamming distance between the inferred and true constraint representations. For binary representations $\widehat{z},z\in\{0,1\}^m$, we compute
\begin{equation}
d_H(\widehat{z},z)
=
\frac{1}{m}
\sum_{j=1}^{m}
\mathbf{1}[\widehat{z}_j\neq z_j].
\label{eq:normalized_hamming}
\end{equation}
For the 2D rod, the representation indicates feasible motion over 72 fixed force directions. For the fixed-base dual-UR5, it indicates which of the 18 possible joint restrictions are active. For the mobile dual-UR5, we compute Hamming distance separately for the base-direction and arm constraints and give the two components equal weight. Lower values indicate more accurate constraint inference.

Zero-shot coordination is evaluated using task success and Success weighted by Number of Actions (SNA). A trial is successful if the robots reach the goal within the setting's action limit. If $n$ is the number of actions executed and $n_{\mathrm{ref}}$ is the reference action count for the same task and physical constraint, we compute
\begin{equation}
\mathrm{SNA}
=
\mathbf{1}[\mathrm{success}]
\frac{n_{\mathrm{ref}}}{\max(n,n_{\mathrm{ref}})}.
\label{eq:sna}
\end{equation}
Failed trials therefore receive an SNA of zero, while successful trials receive higher scores when they complete the task using fewer actions. The reference trajectories use the true physical constraint and are used only for evaluation.

Table~\ref{tab:planner_hyperparameters} summarizes the CEM-MPC parameters used in each setting. All CEM-based methods within a setting use the same planner configuration. The setting-specific CEM hyperparameters were chosen to keep planning runtime approximately comparable across the three environments.

\begin{table}[h]
    \centering
    \caption{CEM-MPC hyperparameters used for zero-shot coordination.}
    \label{tab:planner_hyperparameters}
    \setlength{\tabcolsep}{5pt}
    \begin{tabular}{lrrrr}
        \toprule
        Setting & Horizon & Population & Elites & Iterations\\
        \midrule
        2D Rod & 32 & 64 & 8 & 3 \\
        Fixed-Base Dual-UR5 & 4 & 16 & 2 & 2 \\
        Mobile Dual-UR5 & 4 & 4 & 1 & 1\\
        \bottomrule
    \end{tabular}
\end{table}

Each setting uses three fixed physical-constraint cases for final evaluation. Planning evaluates each constraint on three new tasks with obstacles and one obstacle-free control task using seeds 42, 43, and 44. This gives 27 obstacle-task trials and 9 obstacle-free trials per method and setting. We first average performance across the constraint cases and tasks within each seed, then report the mean and standard error across the three seed-level results. Every scheduled evaluation trial is retained after one attempt, without outcome-dependent filtering or replacement.
\end{document}